# FlexiTokens: Flexible Tokenization for Evolving Language Models

**Abraham Toluwase Owodunni**[1] **Orevaoghene Ahia**[2] **Sachin Kumar**[1]
[1]The Ohio State University [2]University of Washington
owodunni.1@osu.edu

## Abstract

Adapting language models to new data distributions by simple finetuning is challenging. This is due to the rigidity of their subword tokenizers, which typically remain unchanged during adaptation. This inflexibility often leads to inefficient tokenization, causing overfragmentation of text in out-of-distribution domains, unseen languages, or scripts. In this work, we develop byte-level LMs with learnable tokenizers to make tokenization adaptive. Our models include a submodule that learns to predict boundaries given the input byte sequence, encoding it into variable-length segments. Most tokenizer-free methods train this boundary predictor using an auxiliary loss that enforces a fixed compression rate across the training corpus, introducing a new kind of rigidity. We propose FlexiTokens, a simplified training objective that enables significantly greater flexibility during adaptation. Evaluating across multiple multilingual benchmarks, morphologically diverse tasks, and domains, we demonstrate that FlexiTokens consistently reduces token over-fragmentation and achieves up to 10% point improvements on token classification and generative tasks compared to BPE and other gradient-based tokenizers baselines. We validate our findings using models of varying sizes, and our method demonstrates consistent improvements across scales. Code and data for our experiments will be released at https://github.com/skai-research/flexitokens

## 1 Introduction

Tokenization—the process of segmenting text into discrete units—has been shown to significantly influence language model performance (Ali et al., 2024; Geiping et al., 2024; Land and Bartolo, 2024). Widely used subword tokenization algorithms (Sennrich et al., 2016; Devlin et al., 2019) often overfragment sequences in unseen domains, languages, and scripts. This oversegmentation not only leads to poor downstream performance, but also increased sequence lengths, which contribute to higher computational overhead, memory usage, and inference costs (Ahia et al., 2023; Petrov et al., 2023). In addition, such tokenizers are inherently static and tightly coupled with the language model; they do not adapt when the language model is finetuned, e.g., fine-tuning Llama 2 models is subpar for coding tasks (Dagan et al., 2024; Minixhofer et al., 2024), and unseen scripts (Yang et al., 2025). Eliminating the reliance on static subword tokenizers has, thus, gained momentum in recent literature by directly modeling bytes (Xue et al., 2022; Al-Rfou et al., 2018; Wang et al., 2024), although this comes with the overhead of longer sequence length.

To address the increase in sequence length in byte-level language models, various papers introduce a tokenization module within the LM to segment bytes into patches (Tay et al., 2022; Nawrot et al., 2022; Ahia et al., 2024; Pagnoni et al., 2024; Nawrot et al., 2023; Yu et al., 2023) similar to the architecture in Figure 2. As opposed to subword tokenizers, this module is typically learned via gradients alongside the LM with an auxiliary loss to achieve a desired *compression rate* of the input sequence during training. This compression rate, while controllable, is predetermined and fixed during pretraining, which again hampers adaptation to new distributions (see Figure 1). For example, an LM trained with a fixed compression rate on a general domain may over-tokenize samples in specialized domains like Medicine or morphologically rich languages like Turkish that contain longer words. Conversely, it may undertokenize samples in programming languages or logographic languages like Chinese, where distinct semantic units may be inappropriately merged.

To enable flexible adaptation of gradient-based tokenizers, we propose a new training objective, which *relaxes the need to have a fixed* compression rate. Instead, we define a range on the compression rate that every input sequence should have and introduce a hinge-like loss to optimize the tokenization module. By not penalizing the tokenizer when the compression rate is within the range, therefore allowing the predicted compression rate to vary within the range, our method allows for the segmentation to be flexible to the input sequence. When the LM is finetuned, this loss allows the tokenization to adjust to the target distribution without leading to overfragmentation. We call our method FLEXITOKENS (FXT).

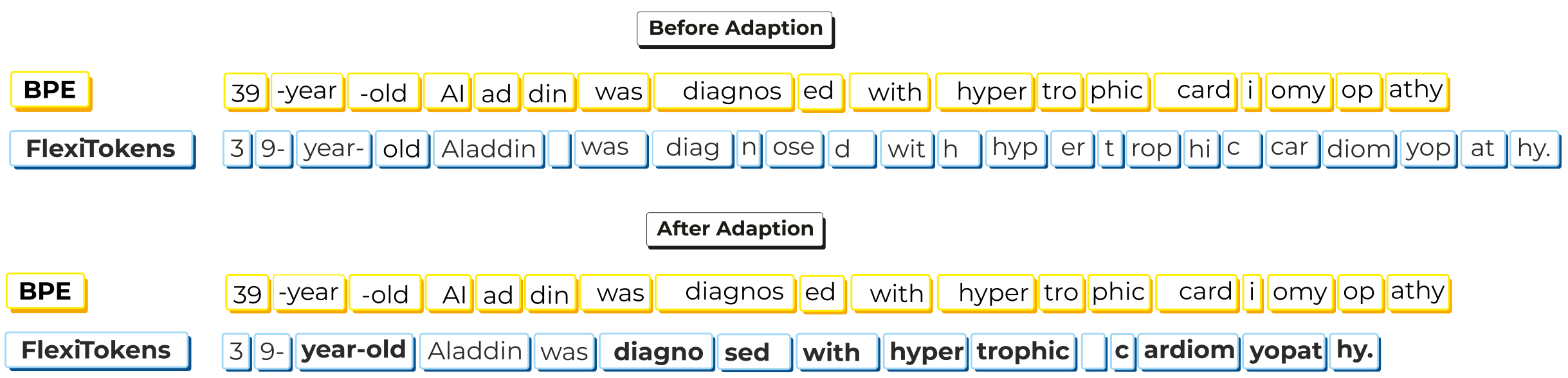

Figure 1: We present an example of tokenized medical text, where FLEXITOKENS produces a less fragmented sequence of tokens than BPE. Unlike BPE, which applies a fixed tokenization, FLEXITOKENS adapts its tokenization to the medical domain, capturing domain-specific patterns more effectively.

We evaluate our proposed approach on multiple multilingual benchmarks and morphologically diverse tasks (§4). FLEXITOKENS consistently shows superior performance compared to baselines while improving average compression rate, thereby improving inference runtime. We also show that while maintaining a fair fragmentation rate across all our pretraining languages, FLEXITOKENS can be easily adapted to unseen languages and scripts without oversegmenting them. Our analysis shows that our method often updates the tokenizer to recover semantically meaningful tokens relevant to the task or domain after adaptation, whereas the baselines, being not updatable, overtokenize.

## 2 FLEXITOKENS

We build a byte-level LM with a learnable tokenization module integrated within the model. FLEXITOKENS allows the model to adjust its learned tokenization strategy to the structure and distribution of the task and input data. Our model builds on the transformer architecture from Nawrot et al. (2023), which is generally used to efficiently handle long sequences in tokenizer-free models (Nawrot et al., 2022, 2023; Ahia et al., 2024; Pagnoni et al., 2024; Hwang et al., 2025). Despite being learnable, the resulting tokenization modules in prior work remain bound to the decisions made during pretraining, even when the model is trained or fine-tuned further. This inherently limits their ability to adapt to new domains, languages, or evolving data distributions, where the originally learned segmentation might no longer be optimal.[1] Below, we first describe the key components of our model adopted from prior works on tokenizer-free models (§2.1) and then introduce the modifications we make to enable dynamic and equitable tokenization (§2.2).

### 2.1 Model Architecture and Overview

Our model, visualized in Figure 2, contains three major components. It consists of a tokenization or downsampling layer that takes byte sequences as input and chunks them into tokens or *patches*. It is then followed by a standard language modeling module, which in turn is followed by an upsampling layer to predict bytes as outputs.

The **tokenization submodule** is a lightweight transformer module (1 to 2 layers) that maps an input byte sequence, $\mathbf{x} = x_1, \ldots, x_N$, to hidden states. Taking these hidden states as input, a boundary predictor then estimates the probability $\hat{b}_t \in [0, 1]$ of predicting a segment boundary at each position $t$.[2] To obtain discrete boundary decisions $b_t \in \{0, 1\}$ while preserving differentiability, a hard Gumbel sigmoid re-parameterization of the Bernoulli distribution is employed. Since this module is differentiable, the segmentations are learned along with the rest of the model.

Given the predicted boundaries, the **language**

[1] This issue is also present in subword tokenizers like BPE. Prior work typically handles this issue with heuristics like retraining and replacing the entire tokenizer during adaptation (Minixhofer et al., 2024).

[2] Different methods of obtaining these probabilities have been employed by prior work. Nawrot et al. (2023) implements it using an MLP (followed by a sigmoid function); Hwang et al. (2025) uses cosine similarity between the query vector of the current position and the key value of the previous position to compute a probability.

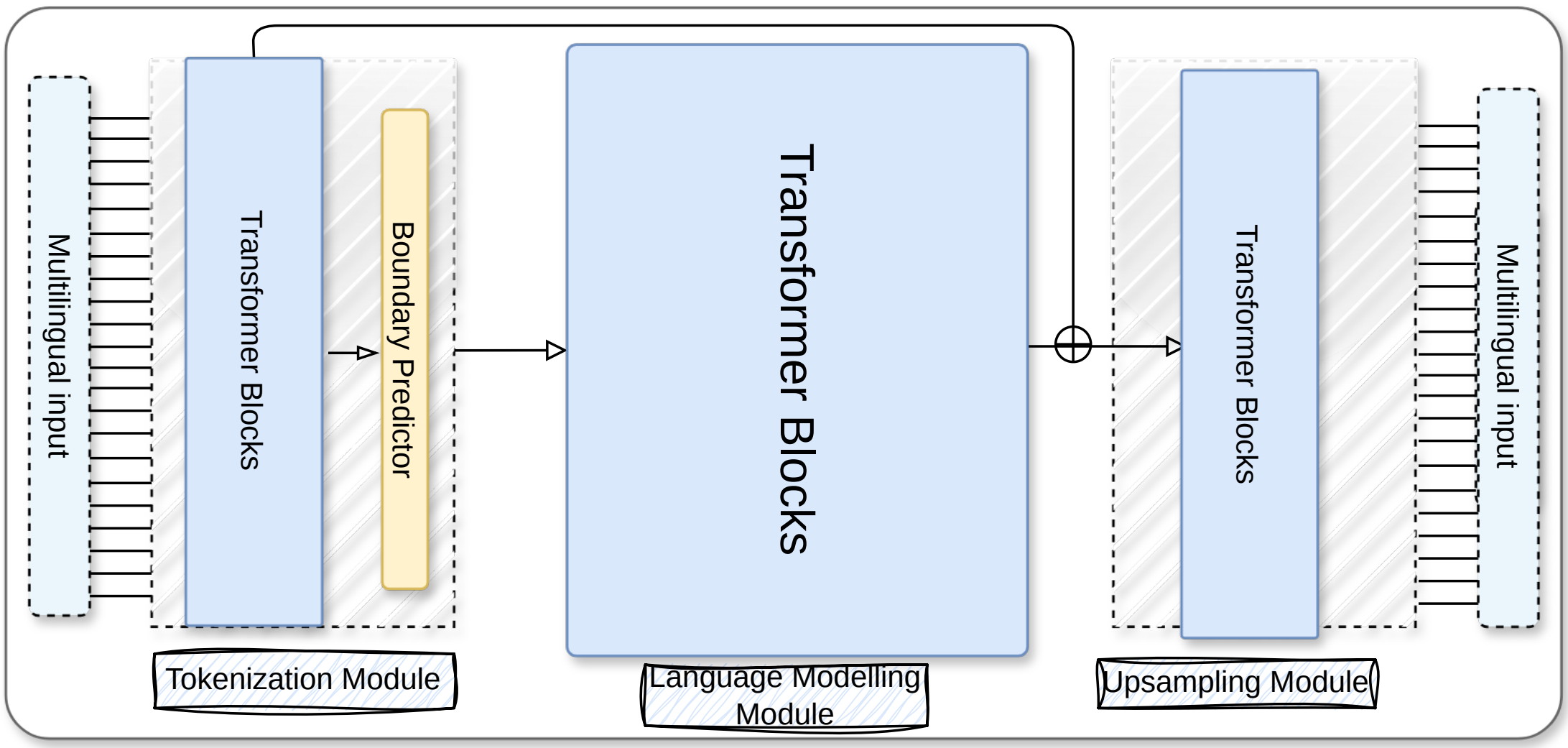

Figure 2: Transformer design used in FLEXITOKENS. The input sequence is passed through the tokenization module for downsampling. The output of the downsampling phase is further processed by the language modeling module and finally upsampled back into the full sequence. Note that all stages use the same transformer block design.

**modeling module** pools the byte-level hidden states between segment boundaries to construct a sequence of token-level representations. These representations are then passed through a stack of transformer layers to obtain another sequence of hidden representations. In a typical LM, this module is equivalent to the transformer blocks without the input and output embedding layers.

Finally, the **upsampling module** converts the outputs from the middle LM block to byte-level probabilities. The token-level representations from the middle block are first upsampled to match the original input resolution via duplication and added to the initial byte-level representations (from the tokenization module) using skip connections. These representations are then passed through another lightweight transformer module (1 to 2 layers), an unembedding layer, and a softmax to produce a probability distribution over the byte vocabulary (of size 256) at each step. The LM loss is computed over the byte vocabulary using these distributions. This modeling backbone is based on the one described in Nawrot et al. (2023); we refer the reader to this paper for a more detailed description.

To prevent the boundary predictor from collapsing and trivially predicting each position $t$ as a boundary, prior work (Nawrot et al., 2023; Ahia et al., 2024) added a regularizer to the LM objective: $-\log \text{Binomial}(\alpha; N, k)$ where,

$$\text{Binomial}(\alpha; N, k) = \binom{N}{k} \alpha^k (1-\alpha)^{N-k} \quad (1)$$

where $k = \sum_{t=1}^{N} b_t$ is the number of boundaries, and $\alpha \in [0, 1]$ is a hyperparameter that controls the expected boundary rate. This loss is lowest when $k$ is close to $\alpha N$, which is the mode of the Binomial distribution. In other words, $\alpha$ controls the compression rate of the input sequence to approximately $\frac{1}{\alpha}\times$. Setting $\alpha = 0$ will cause no boundaries to be predicted, and with $\alpha = 1$, the model learns to predict every position to be a boundary. More recent work has also employed similar boundary-balancing objectives to enforce a compression rate (H-NET; Hwang et al., 2025).

### 2.2 FLEXITOKENS

In contrast with subword-based models like BPE, LMs with gradient-based tokenization can learn to segment input text in a way that best represents the underlying data distribution for language modeling. However, within a language, different subsets, such as different domains, might require different compression rates to optimally encode the input. The expected compression rate, predetermined by $\alpha$, allows little room for variation. Furthermore, when adapting the LM to new distributions, such as a new domain or a new language, bound by the loss in Equation 1, the compression rate does not adapt to the requirements of the target distribution.

The ideal solution to address this issue is to get rid of the hyperparameter $\alpha$ (and the binomial loss) and simply minimize the predicted number of boundaries per byte, that is, $\frac{k}{N}$. However, in our early experiments, we observe that this loss overpowers the optimization process, quickly decreasing to 0, predicting no boundaries. To prevent this behavior, we modify this loss to:

$$\mathcal{L}_{\text{BP}} = \max\left(\frac{k}{N} - \alpha, 0\right) + \max\left(\beta - \frac{k}{N}, 0\right)$$

where $\beta = \alpha - \lambda\sigma$; $\sigma$ represents the standard deviation of tokenization rates over samples in a given language. $\lambda$ is a hyperparameter. This loss introduces a constraint on the boundary rate: $\beta \leq \frac{k}{N} \leq \alpha$. If the boundary rate falls between $\beta$ and $\alpha$, this loss will become 0, reducing further incentive to compress by not penalizing the model. In contrast, losses employed in previous works force the rate to be extremely close to $\alpha$, penalizing both an increase and a decrease at all times. Indeed, we observe in our experiments that there is higher variance in the segmentation rates of different samples. Furthermore, during finetuning, we observe changes in the compression rates, showing that the tokenization indeed adapts to the task. We refer to the flexible tokens learned through our proposed loss and the resulting model that predicts flexible tokens as FLEXITOKENS.[3]

To encode the same information, different languages require different number of bytes, where non-Latin scripted languages (e.g., Indian languages) may require up to 4 bytes per character. When training multilingual models, setting one $\alpha$ for all languages will lead to text in some languages getting segmented into much longer sequences. To alleviate this issue, Ahia et al. (2024) proposed adding a different boundary predictor per language with its own $\alpha$ defined to make the compression rates uniform across languages. A unique boundary predictor per language, however, requires determining or predicting the input language to route the input to the appropriate predictor.

Our experiments reveal that training one shared boundary predictor with a different hyperparameter $\alpha_L$ for each language $L$ leads to the same performance. We train a multilingual model with the following training objective.

$$\sum_{i=1}^{N} - \log p_\theta(x_i \mid x_{<i}) + \sum_{\mathcal{M}} \mathbb{I}(\text{Lang}(\mathbf{x}) = L)\mathcal{L}_{BP_L}$$

where $\mathcal{M}$ is the set of all languages in the train set.

**Determining $\alpha_L$ and $\beta_L$.** We define an anchor language A[4] and set $\alpha_A$ as a hyperparameter. We assume access to a small $n$-way parallel corpus[5] between $A$ and every other language $L$ in our training set.[6] We compute the mean sequence length (in bytes) $\mu_A$, $\mu_L$, and standard deviation $\sigma_A, \sigma_L$ over this dataset. We set $\alpha_L$ to be $\alpha_A \times \frac{\mu_A}{\mu_L}$, and define the lower bound $\beta_L$ as $\alpha_L - \lambda\sigma_L$. Intuitively, if $L$ uses more bytes to represent the same information as $A$, its compression rate should be higher (and hence $\alpha_L$ lower). In summary, only $\alpha_A$ and $\lambda$ are hyperparameters; the others are derived from them.

# 3 Experimental Setup

## 3.1 Datasets

We validate our proposed approach in a multilingual setting. We train models with four scripts and six languages: Latin script (English and Spanish), Cyrillic (Russian and Ukrainian), Devanagari (Hindi), and Telugu script (Telugu). These scripts cover a diverse range of typologies and byte complexities. For example, Latin script needs 1 byte per character in Unicode, whereas Russian and Telugu characters need up to 2 and 3 bytes, respectively. To make segmentation rates similar across all languages, they require different amounts of compression.

For pretraining, we sample a total of 56 billion bytes across multiple languages in FineWeb (Penedo et al., 2024a) and FineWeb 2 (Penedo et al., 2024b). A breakdown of the training set sizes is shown in Figure 6 (in §B). For downstream evaluations, we finetune our models in 9 tasks, including a generative task (translation) and multiple understanding tasks (see Appendix A for more details). In addition, we investigate the robustness of our method by fine-tuning our models on a completely unseen language script that is not included in our pretraining dataset. We also considered evaluating our models on tasks like MMLU (Singh et al., 2024) and INCLUDE (Romanou et al., 2024); however, we obtained close to random chance scores with all baselines and our models due to our smaller scales (see Table 13 in Appendix C).

## 3.2 Hyperparameters

To understand the impact of sequence compression on the model's performance, we explore multiple compression rate configurations. Our main results use 3× compression rate for our anchor language,

[3] We use the term interchangeably to refer to our model and proposed loss.

[4] We choose A as English in all our experiments. This choice is arbitrary; choosing another language will change the $\beta$ values but will not influence the final results.

[5] This computation can also be done with a pairwise parallel dataset with the anchor language with slight modifications.

[6] This parallel dataset is not used for training the model.

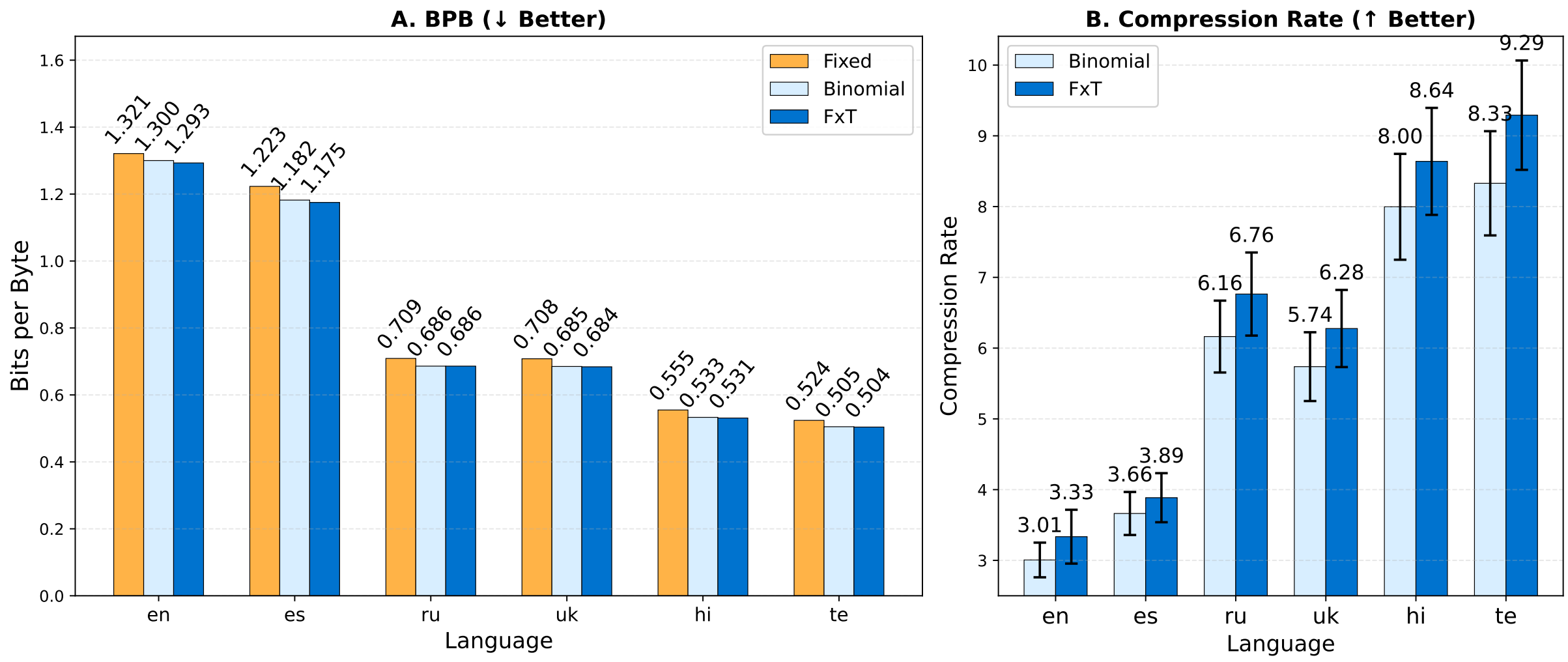

Figure 3: FineWeb Test evaluation results for our medium-sized models. (A.) BPB (↓) of FIXED, BINOMIAL and FLEXITOKENS (FXT). (B.) Compression rate (↑) and Compression variance (error bars) of FLEXITOKENS compared to the BINOMIAL variant with $\alpha_A = 0.3$ and $\lambda = 3$. FIXED model has a fixed compression rate and variance, which can be seen in Table 5. Higher compression rates result in fewer tokens and more compounded tokens, which in turn leads to a more efficient model. FLEXITOKENS outperforms all baselines on BPB while also achieving higher sequence compression and variance.

English (i.e. $\alpha = 1/3$). We also compare with 5× and 10× compression rates. The corresponding values of $\alpha_L$ and $\sigma_L$ for all languages are in Appendix Table 5. We compute the lower bound $\beta_L$ using the FLORES-200 dataset (Costa-Jussà et al., 2022), which contains parallel sentences in 200 languages. We note that these values can be computed using any pairwise parallel corpus, which we demonstrate in Table 6. We empirically set $\lambda = 3$; we show comparisons with other values in Table 3. In our experiment with adapting our model to an unseen script (for Urdu), we set $\beta$ to have the same value as Telugu, which has the highest compression rate of all the languages we experimented on, assuming no available parallel dataset in the unseen language.

**Model Architecture and Pretraining** We pretrain two model sizes: FXT-SMALL (126M parameters, comparable to GPT2 small) and FXT-medium (388M parameters, comparable to GPT2 medium). For SMALL model, we follow Ahia et al. (2024) to create a 16-layer transformer model with 2 transformer layers in the tokenization and upsampling modules each. The boundary predictor is a 2-layer MLP. See Appendix B for specific details of our model architecture.

### 3.3 Baselines

We compare with 3 strong baselines, which include a model trained with a BPE tokenizer, a model with static byte pooling (where every $k$ tokens are pooled together to obtain a $k\times$ compression), and a byte-level model whose boundary predictor is trained with a binomial loss as described in Nawrot et al. (2023) (BINOMIAL). We note that our method can also be applied to any model that uses an extra loss to optimize its boundary predictor. For fair comparison with the BPE model, we match its overall parameter size with FLEXITOKENS. We train a BPE tokenizer with a vocabulary size of 50K on the same amount of dataset from each language; it achieves a compression rate of 4.4× on English.[7] To match total parameters (embeddings + transformer layers), we train the LM with 5 and 20 Transformer layers for our small and medium-sized BPE models, respectively.[8]

## 4 Results and Analyses

We evaluate our pretrained model using bits per byte (BPB) (Graves, 2013) and the finetuned models using task-specific metrics, mostly accuracy and F1-score. We provide a summary of the results for the pretrained models in Figure 3 and Figure 4, and for our finetuned small and medium-sized models in Tables 1, 2, 15, and Figure 5, with details in Ap-pendix C. We perform our analyses in this section with models trained with $\lambda = 3$, which we find produce the best result in most cases.

[7] Note that BPE models cannot be controlled to have desired compression rates across all languages due to their inherent frequency-based training process (Ahia et al., 2023).

[8] We conducted early experiments with training BPE-based models by matching English's compression rate to 3× compression rate, but they resulted in vocabulary sizes of 10K, which performed poorly.

**Pretraining with FLEXITOKENS leads to better compression** As shown in Figure 3, our method achieves the best BPB performance compared to other byte models on the FineWeb test sets while achieving a substantially higher average compression rate, which in turn increases inference speed by requiring fewer tokens. The BPB metric is the byte-normalized cross-entropy (in bits) over a given text. It is the log-space analogue of perplexity and used for tokenizer-independent evaluations. From our experiments, we do not observe any strong correlation between BPB and compression rate. Furthermore, we extend our experiments to compare with our boundary loss in FLEXITOKENS with the loss proposed in HNet (Hwang et al., 2025). We find that (Table 14 in Appendix C.1) FLEXITOKENS maintains similar BPB as HNet while achieving a higher compression rate.

We also observe a higher variance in compression rates of FLEXITOKENS (Figure 3), implying higher flexibility in how input sequences are fragmented. This variation—which is much lower in baseline models—alongside the higher compression rate on average underscores FLEXITOKENS' ability to dynamically adapt its tokenization patterns to its input. In Figure 4, we compare the average number of tokens required to represent the same information in different languages by different tokenization methods. Our method remains as equitable as BINOMIAL using a similar number of tokens for all languages. In comparison, BPE shows high variability with included languages like Hindi and Telugu requiring twice as many tokens. An unseen language (Urdu) requires 6× as much.

**FLEXITOKENS adapts tokenization and boosts performance across tasks and domains.** In Tables 2, 10 and 15, we report task-specific metrics after finetuning our pretrained models on several downstream tasks across different domains and the corresponding compression rates per language and task in Figure 5. FLEXITOKENS outperforms all baselines on the majority of tasks, including the BPE baseline, with a much higher compression rate. Our method obtains performance improvements of over 3 points on XNLI compared to BINOMIAL while improving compression across all tasks. Notably, we observe up to 21 points of improvement over BPE on our generative task (translation) shown in Table 1. We also notice that as we increase $\lambda$, performance tends to also increase. This is because a higher $\lambda$ allows a wider margin for the model to find the optimal tokenization pattern.

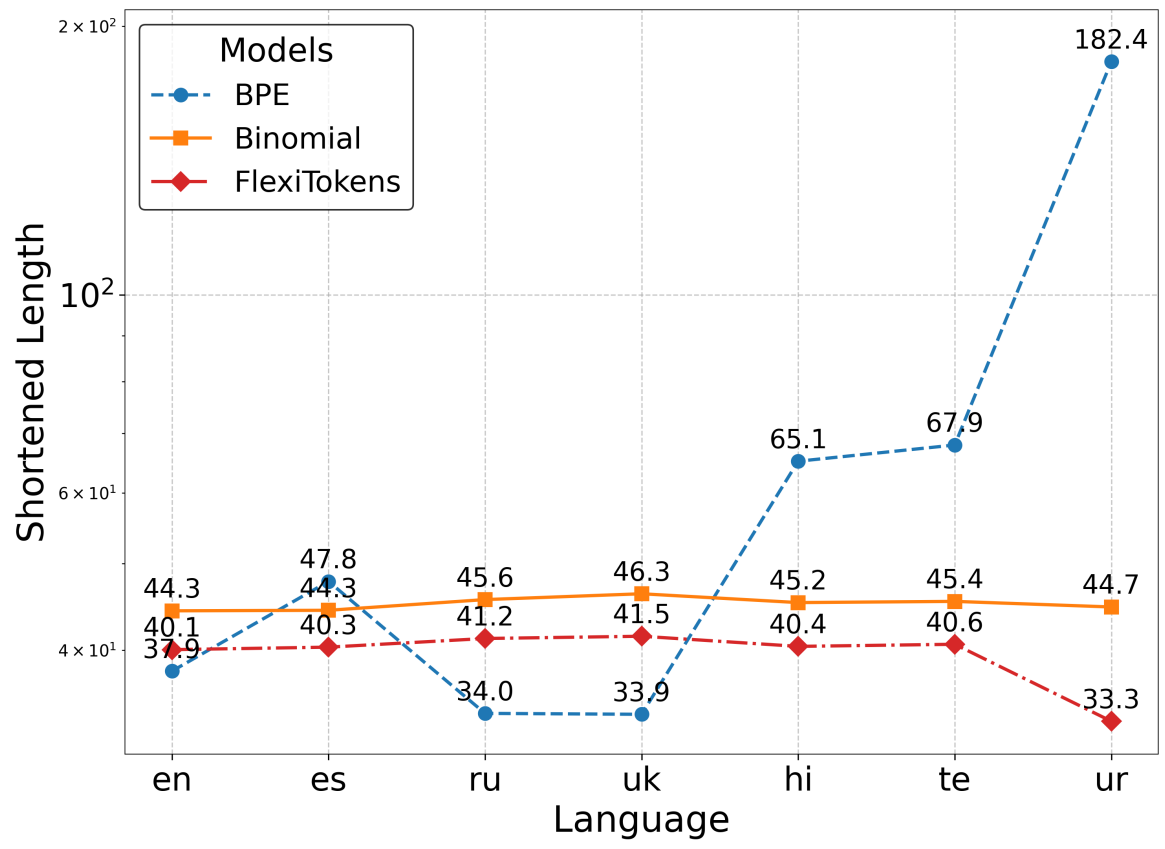

Figure 4: Average number of tokens per sample obtained in the FLORES dataset with different tokenization algorithms. FLEXITOKENS consistently produces the least number of tokens while maintaining balance across languages, even for the unseen language Urdu. BPE over-fragments seen (Hindi, Telugu) as well as unseen languages (Urdu).

Analyzing compression rates across tasks and languages in Figure 5, we observe that BINOMIAL tends to maintain rates closer to the initial $\alpha$, but this effect diminishes for non-Latin languages such as Hindi and Telugu, which are structurally distant from Latin scripts. These languages show both higher average compression and greater variance with FLEXITOKENS.

Qualitative analysis in Figure 5 reveals consistent tokenization patterns across topic classification tasks like SIB-200 and WikiANN NER, where compression remains relatively stable across examples. In contrast, tasks such as XNLI exhibit compression spikes across all languages, indicating that some tasks benefit from more tokenization boundary allowance than others. In the Irony Classification task, FLEXITOKENS effectively tokenizes emojis with higher compression, preserving their semantic meaning. Following adaptation to the medical domain (Figure 1), we also find that medical terms are tokenized in unison as whole words (Figure 1, reducing fragmentation and better aligning with expected domain-specific vocabulary.

**Adaptive tokenization to unseen scripts boosts performance without overfragmentation** In Table 2, we extend our evaluation to Urdu, a low-resource Indo-Aryan language that shares linguis-tic commonalities with Hindi but uses a different script, not included in our pretraining dataset. We see that FLEXITOKENS outperforms BPE with more than 3 points after finetuning. Qualitative evaluation on the XNLI inputs (Table 21) reveals that our approach finds more compressed and semantically meaningful tokens compared to baselines (numbers and words). BPE tokenizer tokenizes Urdu with more 6× tokens than FLEXITOKENS which follows the same result patterns from Figure 4. Note that FLEXITOKENS adapts well to unseen scripts because we use a script-agnostic boundary predictor as opposed to Ahia et al. (2024), which introduced the idea of equitable tokenization via a script-specific boundary predictor for every language script included during pretraining. Also, compound or rare words (especially medical terms or foreign-origin words like "hypertrophic") are split into meaningful subwords, enabling the model to learn more meaningful representations (Figure 1).

| Model | en-es | en-ru | en-hi |
|---|---|---|---|
| BPE | 68.93 | 62.31 | 49.98 |
| FIXED | 70.75 | 54.28 | 58.56 |
| BINOMIAL | 67.65 | 61.01 | 60.44 |
| FXT $\lambda 3$ | **71.11** | **62.88** | **61.46** |

Table 1: COMET scores on OPUS-100 machine translation task for our **medium-sized** models. FLEXITOKENS outperforms across all language pairs. See Table 15 for the results of our small models.

**Impact of scaling each module:** We experimented with scaling FLEXITOKENS by adding more layers to the tokenization, language modeling, and upsampling module. Overall, we observe (see Figure 7 in Appendix D) that increasing our model's parameters by adding more layers to each module improves performance. We also note that the compression rate increases as we add more layers to the model. We speculate that this gain is because more layers allow the model to create richer representations prior to tokenization. We note that for the choice of which module gains the most from layer addition, increasing the LM module with 2 layers (2,14,2) outperforms adding more layers to other parts of the model (3,12,3). These results provide insightful directions for future research on scaling FLEXITOKENS.

**Relationship between compression and model performance:** We explore various configurations of $\alpha$ and how it impacts performance and show average results across all tasks in Table 3 (see Appendix D for a breakdown of performance on each language). As we scale the compression rate from 3× to 5× and 10×, we observe a slight decline in performance as models use less training FLOPS, indicating that too much forced compression may result in loss of information, hurting the model. We also speculate that this issue might be attributed to the small model size used in our main experiments. Recent work has argued that larger models can handle larger vocabularies better (Tao et al., 2024). Its analogue in our case is training a larger model with more layers in the tokenization module, which we show improves performance in FLEXITOKENS (3,12,3). Further scaling of both the tokenization and LM module is likely to further improve performance.

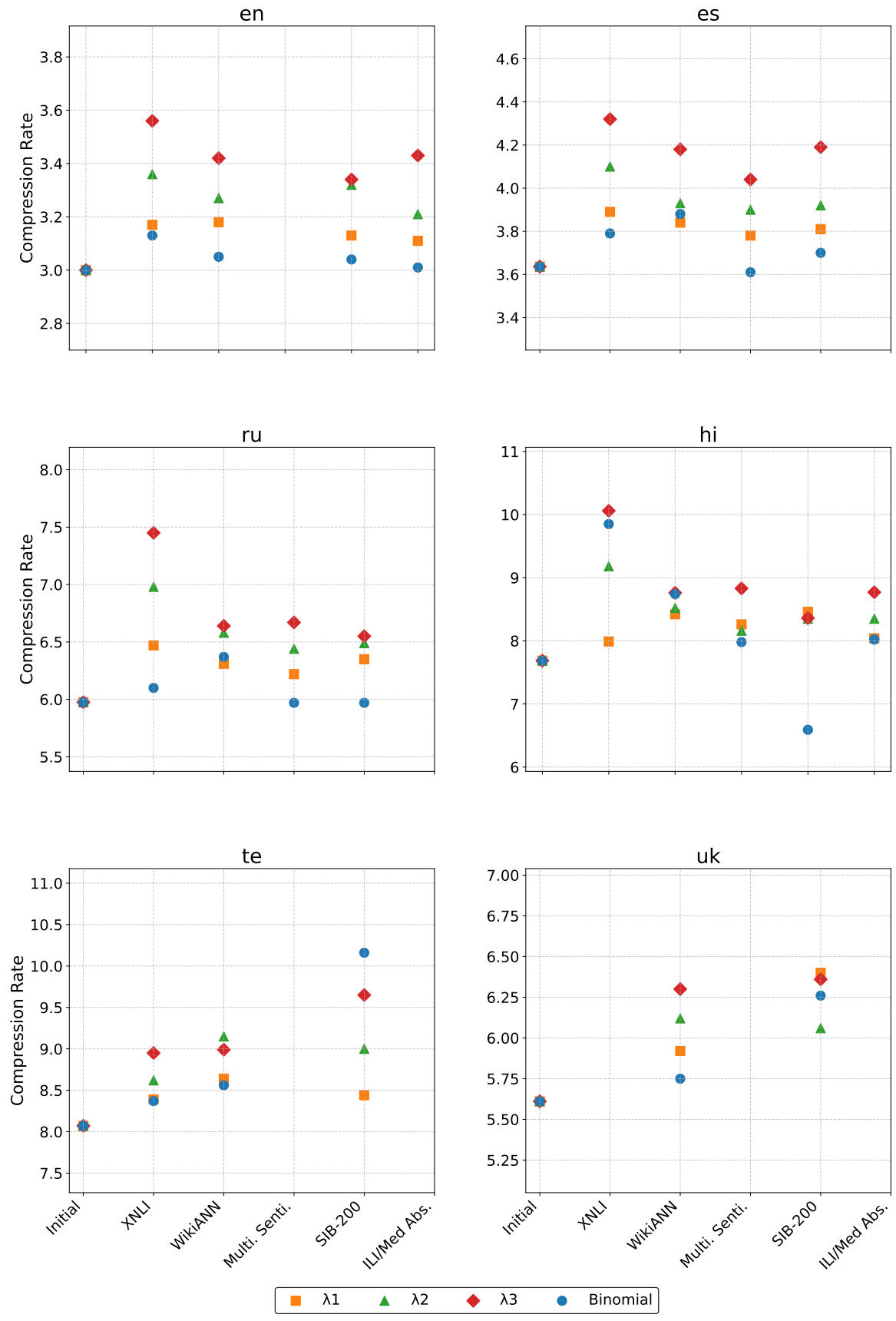

Figure 5: Compression rate changes with FLEXITOKENS across multiple tasks. *Initial* is the base compression rate before pretraining. Compression rate for BINOMIAL remains relatively low while we also see a spike for tasks like XNLI.

## 5 Related Work

**Tokenizer-free language modeling** Several works have explored the possibilities of training language models without relying on subword tokenization, instead representing text directly as a sequence of bytes (Xue et al., 2022; Al-Rfou et al., 2018; Wang et al., 2024; Limisiewicz et al., 2024) or pixels (Lotz et al., 2023; Rust et al., 2023; Salesky et al., 2023). To address the efficiency challenges of processing raw characters or byte sequences on tokenizer-free LMs, alternative architectures have proposed to either segment byte sequences into fixed-length (Nawrot et al., 2022; Clark et al., 2022; Godey et al., 2022; Tay et al., 2022; Yu et al., 2023) or dynamic segments (Nawrot et al., 2023; Deiseroth et al., 2024; Ahia et al., 2024; Pagnoni et al., 2024; Hwang et al., 2025). However, these models are pretrained with a fixed target compression rate, which limits their ability to adapt to shifts in data distribution. Moreso, the H-Net hierarchical transformer architecture (Hwang et al., 2025) has been found by prior works to be subpar to BPE and unstable during training (Main, 2025).

| NER F1 Score | | | | | | | |
|---|---|---|---|---|---|---|---|
| **Model** | **en** | **es** | **ru** | **uk** | **hi** | **te** | **Avg** |
| BPE | 52.30 | 67.70 | 64.94 | 74.99 | 60.23 | 48.18 | 61.39 |
| FIXED | 63.56 | 76.15 | 68.34 | 78.21 | 63.01 | 51.47 | 66.79 |
| BINOMIAL | 63.80 | 75.06 | 67.59 | **78.06** | 61.21 | 48.31 | 65.67 |
| FXT $\lambda 1$ | 63.07 | 76.12 | **68.30** | 77.94 | **62.26** | 51.74 | **66.57** |
| FXT $\lambda 2$ | **63.96** | **76.23** | 67.55 | 77.99 | 62.24 | 48.13 | 66.02 |
| FXT $\lambda 3$ | 63.73 | 75.45 | 68.25 | 78.01 | 61.97 | 52.02 | 66.55 |

| XNLI Accuracy | | | | | | | |
|---|---|---|---|---|---|---|---|
| **Model** | **en** | **es** | **ru** | **hi** | **te** | **ur** (OOD) | **Avg** |
| BPE | 73.09 | 69.9 | 65.95 | 61.48 | **68.00** | 54.11 | 65.42 |
| FIXED | 73.25 | 69.84 | 66.47 | 61.59 | 67.52 | 57.09 | 65.96 |
| BINOMIAL | 72.87 | 70.28 | 65.93 | 62.26 | 66.11 | 54.79 | 65.37 |
| FXT $\lambda 1$ | **73.51** | 70.22 | 66.47 | **62.42** | 67.11 | 56.99 | 66.12 |
| FXT $\lambda 2$ | 73.21 | **70.84** | **66.97** | 62.16 | 66.71 | **57.58** | 66.25 |
| FXT $\lambda 3$ | 73.35 | 70.22 | 66.75 | 62.36 | 67.82 | 57.33 | **66.31** |

| Med. Abs./Irony/CS/CS/ILI - Accuracy | | | | | | | |
|---|---|---|---|---|---|---|---|
| **Model** | **Med. Abs.** **en** | **Irony** **en** | **CS** **en-es** | **CS** **en-hi** | **ILI** **hi** | **-** **-** | **Avg** |
| BPE | 57.68 | 67.86 | **92.48** | **87.36** | 89.06 | - | 78.89 |
| FIXED | 60.18 | 65.68 | 91.86 | 86.17 | 89.03 | - | 78.58 |
| BINOMIAL | 62.81 | 67.60 | 91.62 | 84.98 | 89.47 | - | 79.30 |
| FXT $\lambda 1$ | 62.92 | 68.37 | - | - | 89.58 | - | 73.62 |
| FXT $\lambda 2$ | 62.74 | 68.75 | 91.37 | 86.53 | **90.33** | - | 79.94 |
| FXT $\lambda 3$ | **63.19** | **69.26** | 92.11 | 86.41 | 89.55 | - | **80.10** |

Table 2: WikiANN (NER), XNLI and other tasks' F1 Score and Accuracy and for 3× Compression Rate on our small-sized models. FLEXITOKENS outperforms all baselines on XNLI and NER respectively. Notably, it achieves approximately a 3-point gain on XNLI for Urdu—an unseen language script—compared to BPE.

**Adapting tokenizers to new distributions** In adapting tokenizer-free LMs to new data distributions, Mofijul Islam et al. (2022) propose a character-based tokenizer by distilling segmentation information from heuristic-based subword tokenization. In contrast, several studies have explored adaptation strategies for subword tokenizers, both at inference time and during fine-tuning. For instance, prior work has shown that improved segmentation of large numbers can enhance performance on arithmetic tasks without retraining (Singh and Strouse, 2024; Sathe et al., 2025). (Godey et al., 2022) also studied adapting gradient-based models to new distribution, however their work only provides limited experimental evidence at small scale. In multilingual and domain-specific settings, various approaches have been proposed to adapt subword tokenizers during fine-tuning. These involve refining the tokenizer vocabulary with new tokens from the target distribution and initializing the corresponding embeddings to better capture linguistic and domain-specific characteristics (Park et al., 2021; Alabi et al., 2022; Minixhofer et al., 2022; Sachidananda et al., 2021; Liu et al., 2023). However, our experiments indicate that subword tokenizers often underperform in low-resource and non-Latin script languages due to over-segmentation. Recent work has also explored converting models trained with subword tokenizers into tokenizer-free models (Minixhofer et al., 2025), albeit, the resulting tokenizer-free models consistently performed subpar to their subword counterparts.

| Model | SIB-200 | WikiANN | Multi. Senti. | XNLI | ILI | Med. Abs. | Avg |
|---|---|---|---|---|---|---|---|
| | | | **Accuracy** | | | | |
| FxT 10x | 53.76 | 64.35 | **72.99** | 65.23 | 89.07 | 62.95 | 68.06 |
| FxT 5x | 71.16 | 64.92 | 72.54 | 65.48 | 89.28 | **63.47** | 71.14 |
| FxT 3x | **72.55** | **66.02** | 72.74 | **66.25** | **90.33** | 62.74 | **71.77** |
| | | | **Compression Rate ± Std** | | | | |
| FxT 10x | 28.89 ± 11.06 | 28.01 ± 14.14 | 27.41 ± 12.12 | 29.06 ± 8.55 | 38.80 ± 38.80 | 13.22 ± 2.15 | 27.56 ± 14.47 |
| FxT 5x | 10.72 ± 1.54 | 11.17 ± 3.69 | 11.25 ± 2.86 | 12.15 ± 1.76 | 14.82 ± 14.82 | 5.63 ± 0.33 | 10.96 ± 4.17 |
| FxT 3x | 6.19 ± 0.53 | 6.26 ± 1.33 | 6.17 ± 1.03 | 6.83 ± 0.60 | 8.35 ± 8.35 | 3.21 ± 0.15 | 6.17 ± 2.00 |

Table 3: Ablation for $\alpha$: Average Accuracy and Compression Results Across Multiple Languages

## 6 Conclusion

We introduced FLEXITOKENS, a flexible, gradient-based tokenization approach that enables language models to adapt their segmentation patterns during finetuning. Unlike prior methods that enforce static or fixed compression rates, our method promotes dynamic tokenization aligned with the structure of the target distribution. Through multilingual and domain-diverse evaluations, FLEXITOKENS consistently reduces token over-fragmentation, improves downstream task performance, and achieves higher compression without sacrificing accuracy. Our results highlight the importance of adaptable tokenization strategies for building more efficient and generalizable language models.

## Limitations

Our limited computational budget prevents us from training larger models with more language on more datasets. We anticipate the results will improve with scaling, potentially providing even higher compression. We leave this exploration to future work. While we aimed for diversity of languages and scripts in our experiments, we acknowledge that we do not cover a vast majority of linguistic diversity. But our methods are general, and we believe our results should translate to more languages. We also acknowledge a tradeoff between the performance and compression rate of the languages, with higher compression leading to a slight decline in performance, with some languages being more sensitive than others. FLEXITOKENS shares limitations of other segmentation methods in that it may not be suitable for languages where morphemes are discontinuous and vowels are interspersed between consonant roots for inflection or sometimes omitted, such as Semitic languages or other languages with Templatic morphologies.

## Appendix

## A Dataset

Here, we provide more details of how we create the pertaining dataset for our finetuning tasks. In our pretraining corpus, we sample the first 2.06M documents from FineWeb (Penedo et al., 2024a) for English, using the first 10K documents as the validation set. For all other languages, we sample the first 1.65M documents from FineWeb 2 (Penedo et al., 2024b) (see Figure 6), again using the first 10K documents for validation.

| **Dataset** | **en** | **es** | **ru** | **uk** | **hi** | **te** | **ur** |
|---|---|---|---|---|---|---|---|
| XNLI | 64 | 64 | 64 | 64 | 64 | 64 | 64 |
| SIB-200 | 8 | 8 | 8 | 8 | 8 | 8 | - |
| WikiANN | 16 | 16 | 16 | 16 | 16 | 16 | - |
| Multi. Sentiment | 128 | 32 | 32 | - | 8 | - | - |
| Machine Transl. | - | 64 | 64 | - | 64 | - | - |
| Code-Switch | - | 32 | - | - | 32 | - | - |
| ILI | - | - | - | - | 32 | - | - |
| Medical Abstract | 16 | - | - | - | - | - | - |
| Irony detection | 32 | - | - | - | - | - | - |

Table 4: Batch Sizes per Dataset and Language

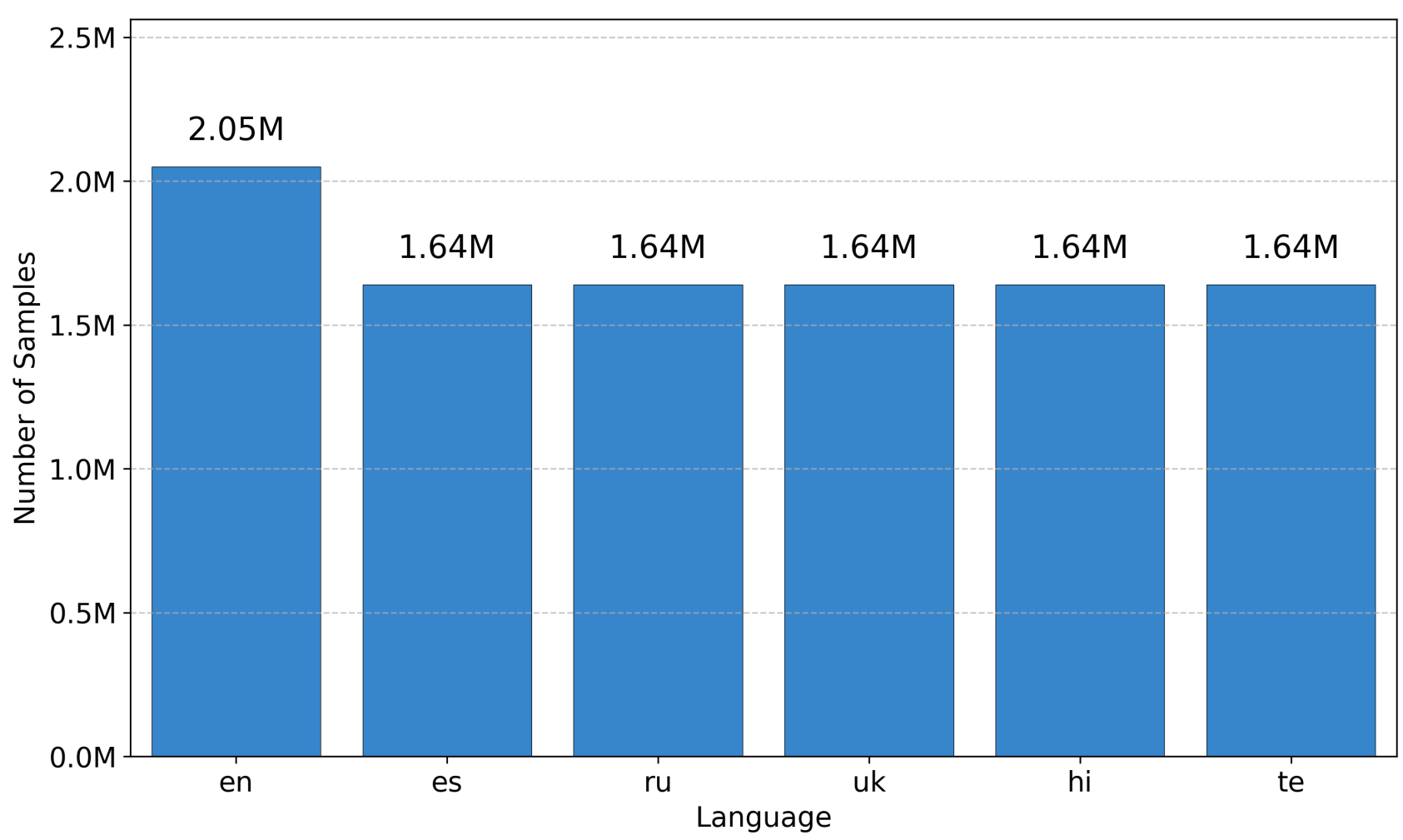

Figure 6: Number of training documents sampled by language

To evaluate our method, we finetune our model on the following tasks: (1) *OPUS-100* (Tiedemann, 2012): machine translation, (2) *XNLI* (Conneau et al., 2018): natural language inference, (3) *SIB-200* (Adelani et al., 2023): topic classification, (4) *Multilingual Sentiment* (clapAI, 2024): multi-domain sentiment analysis, (5) *WikiANN* (Pan et al., 2017): named entity recognition, (6) *Indo-Aryan Language Identification (ILI)*[9] (Zampieri et al., 2018): dialect classification, (7) *SwitchLingua* (CS) (Zampieri et al., 2018): code-switched text classification; (8) *Medical Abstracts Text Classification* (Schopf et al., 2022) and (9) *Irony detection* in Tweets containing emojis (Rohanian et al., 2018).

[9]https://github.com/kmi-linguistics/vardial2018

| Configuration | en | es | ru | uk | hi | te |
|---|---|---|---|---|---|---|
| FxT 10× | 0.1 / 10 | 0.08 / 12.12 | 0.05 / 19.92 | 0.053 / 18.70 | 0.039 / 25.62 | 0.037 / 26.91 |
| FxT 5× | 0.2 / 5 | 0.17 / 6.06 | 0.1 / 9.96 | 0.107 / 9.35 | 0.078 / 12.81 | 0.074 / 13.45 |
| FxT 3× | 0.333 / 3 | 0.28 / 3.64 | 0.167 / 5.98 | 0.178 / 5.61 | 0.13 / 7.68 | 0.124 / 8.07 |
| $\sigma$ | 0.023 | 0.019 | 0.011 | 0.012 | 0.009 | 0.008 |

Table 5: $\alpha_L$ and $\sigma_L$ values for each language in our training dataset, computed using FLORES-200.

## B Model Architecture and Training Hyperparameters

In Figure 2, we present the FLEXITOKENS transformer design used in this work. We also show the flow of an input sequence through all the modules in the model. All other parameters follow Ahia et al. (2024), except for the boundary predictor: instead of multiple predictors, we use a single 2-layer MLP as the boundary predictor. Our full list of hyperparameters is highlighted in Table 7. We use the implementation by Nawrot et al. (2022), since their training code is publicly available and it has been successfully adopted by other works (Fleshman and Van Durme, 2023; Ahia et al., 2024).

We present the precomputed values of $\alpha_L$ and $\sigma_L$ in Table 5 using the FLORES-200 dataset. We also explored using other datasets to precompute the values of $\alpha_L$ in Table 6 to ensure consistency across languages, and we find these numbers to be consistently similar.

| Language | FLORES-200 | MGSM |
|---|---|---|
| en | 0.356 | 0.363 |
| es | 0.294 | 0.332 |
| ru | 0.178 | 0.199 |
| te | 0.132 | 0.134 |

Table 6: Precomputed $\alpha_L$ values using FLORES-200 and MGSM datasets.

**Training Configurations and Hyperparameters** For pretraining, we use sequence lengths of 512 and 2048 bytes in our small and medium-sized models, respectively. We train across 4 H100 GPUs using the hyperparameters in Table 7. During finetuning, we increase all models' sequence lengths to 2048 bytes to better capture longer sequences in the finetuning dataset.[10] For the NER task, we first concatenate token sequences using whitespaces before tokenization and label whitespaces as non-entity. All tasks are finetuned for 5 epochs and an LR of 5e-5, except for OPUS-100 for which we finetuned for 2 epochs LR of 5e-4. We use task-specific batch sizes based on data availability. We perform monolingual finetuning on each language. Please refer to Table 4 for full finetuning parameters.

## C Results and Analyses

In this section, we present the full results discussed in §4 across all our selected downstream tasks as seen in Table 9, 10, 11, and 12. We also present the full results for our multilingual sentiment analysis (Table 8) and SIB-200 (Table 10) evaluations . All Results in this section contain values for performance metrics like accuracy and F1 score, compression rates, and standard deviation of the compression rates.

### C.1 Comparison with H-Net

Here we also report results of comparing our method with HNet by training multiple HNet (Hwang et al., 2025) models with 3x compression rate using the original HNet boundary predictor loss and our boundary predictor loss in FLEXITOKENS. In Table 14, we find that FLEXITOKENS maintains the same BPB as HNet, while achieving an even higher compression rate.

[10]We use a shorter sequence length during pretraining due to computational constraints.

| **Hyperparameter** | **FxT-Small** | **FxT-Medium** |
|---|---|---|
| *Model Architecture* | | |
| Parameters | 126M | 388M |
| Model Dimension ($d_{\text{model}}$) | 768 | 1024 |
| Attention Heads ($n_{\text{head}}$) | 12 | 16 |
| Head Dimension ($d_{\text{head}}$) | 64 | 64 |
| FFN Inner Dimension ($d_{\text{inner}}$) | 3072 | 4096 |
| Sequence Length | 512 | 2048 |
| Tokenization Module Layers | 2 | 2 |
| LM Module Layers | 12 | 24 |
| Upsampling Module Layers | 2 | 2 |
| Dropout | 0.1 | 0.1 |
| Attention Dropout | 0.1 | 0.1 |
| *Optimization* | | |
| Optimizer | Adam | Adam |
| Learning Rate | 5e-5 | 1e-4 |
| LR Scheduler | Cosine | Cosine |
| Warmup Steps | 9000 | 6000 |
| Max Training Steps | 100000 | - |
| Batch Size | 512 | 256 |
| Gradient Clipping | 0.25 | 0.25 |
| Weight Decay | 0 | 0 |
| Adam $\beta_1$ | 0.9 | 0.9 |
| Adam $\beta_2$ | 0.999 | 0.999 |
| Adam $\epsilon$ | 1e-8 | 1e-8 |

Table 7: Hyperparameters for FxT-Small and FxT-Medium

## D Full Ablation Results

We present the full ablation results as discussed in §4 in Table 3. All results in this section (17, 18, 19, 20, and 16) contain values for performance metrics like accuracy and F1 score, compression rates, and standard deviation of the compression rates. We also present the results of our ablation study on how scaling the model's tokenization module impacts performance in Figure 7

| Model | es | ru | hi | Avg |
|---|---|---|---|---|
| | | **Accuracy** | | |
| BPE | – | – | – | – |
| Binomial 3x | **77.89** | 87.20 | **53.63** | **72.91** |
| FxT $\lambda$1 | 77.75 | **87.33** | 53.42 | 72.83 |
| FxT $\lambda$2 | 77.77 | **87.33** | 53.12 | 72.74 |
| FxT $\lambda$3 | 77.63 | 87.13 | 53.01 | 72.59 |
| | | **Compression Rate ± Std** | | |
| Binomial | 3.61 ± 0.48 | 5.97 ± 0.98 | 7.98 ± 1.90 | 5.85 ± 1.27 |
| FxT $\lambda$1 | 3.78 ± 0.27 | 6.22 ± 0.53 | 8.26 ± 1.82 | 6.09 ± 1.11 |
| FxT $\lambda$2 | 3.90 ± 0.28 | 6.44 ± 0.61 | 8.16 ± 1.65 | 6.17 ± 1.03 |
| FxT $\lambda$3 | 4.04 ± 0.37 | 6.67 ± 0.75 | 8.83 ± 1.84 | 6.51 ± 1.17 |

Table 8: Multilingual Sentiment Accuracy and Compression Results for 3x Configurations

| Model | en | es | ru | uk | hi | te | Avg |
|---|---|---|---|---|---|---|---|
| | | | | **F1 Score** | | | |
| BPE | 52.30 | 67.7 0 | 64.94 | 74.99 | 60.23 | 48.18 | 61.39 |
| FIXED | 63.56 | 76.15 | 68.34 | 78.21 | 63.01 | 51.47 | 66.79 |
| Binomial | 63.80 | 75.06 | 67.59 | **78.06** | 61.21 | 48.31 | 65.67 |
| FxT $\lambda$1 | 63.07 | 76.12 | **68.30** | 77.94 | **62.26** | **51.74** | **66.57** |
| FxT $\lambda$2 | **63.96** | **76.23** | 67.55 | 77.99 | 62.24 | 48.13 | 66.02 |
| FxT $\lambda$3 | 63.73 | 75.45 | 68.25 | 78.01 | 61.97 | 52.02 | 66.38 |
| | | | | **Compression Rate ± Std** | | | |
| Binomial 3x | 3.05 ± 0.47 | 3.88 ± 0.76 | 6.37 ± 1.67 | 5.75 ± 1.11 | 8.74 ± 3.27 | 8.56 ± 2.29 | 6.06 ± 1.86 |
| FxT $\lambda$1 | 3.18 ± 0.43 | 3.84 ± 0.54 | 6.31 ± 1.15 | 5.92 ± 0.90 | 8.42 ± 1.68 | 8.64 ± 1.55 | 6.05 ± 1.14 |
| FxT $\lambda$2 | 3.27 ± 0.44 | 3.93 ± 0.58 | 6.58 ± 1.38 | 6.12 ± 1.00 | 8.52 ± 1.49 | 9.15 ± 2.21 | 5.66 ± 1.33 |
| FxT $\lambda$3 | 3.42 ± 0.53 | 4.18 ± 0.66 | 6.64 ± 1.29 | 6.30 ± 1.07 | 8.76 ± 1.77 | 8.99 ± 2.07 | 6.38 ± 1.35 |

Table 9: WikiANN NER F1 Score and Compression Results for 3x Configurations

| Model | en | es | ru | uk | hi | te | Avg |
|---|---|---|---|---|---|---|---|
| | | | | **Accuracy** | | | |
| BPE | **80.88** | **81.37** | **81.37** | **76.96** | 60.78 | **72.55** | **75.65** |
| FIXED | 76.47 | 74.51 | 76.47 | 68.63 | 63.23 | 69.61 | 71.49 |
| Binomial | 79.41 | 74.02 | 71.08 | 68.63 | 64.71 | 69.61 | 71.24 |
| FxT $\lambda$1 | 78.92 | 72.55 | 75.49 | 69.61 | 61.27 | 66.18 | 70.67 |
| FxT $\lambda$2 | 77.94 | 75.98 | 74.51 | 71.57 | 69.12 | 66.18 | 72.55 |
| FxT $\lambda$3 | **80.88** | 77.45 | 73.04 | 72.55 | **71.08** | **71.08** | 74.35 |
| | | | | **Compression Rate ± Std** | | | |
| Binomial | 3.04 ± 0.27 | 3.70 ± 0.34 | 5.97 ± 0.64 | 6.26 ± 0.70 | 6.59 ± 0.48 | 10.16 ± 1.34 | 5.95 ± 0.72 |
| FxT $\lambda$1 | 3.13 ± 0.25 | 3.81 ± 0.29 | 6.35 ± 0.64 | 6.40 ± 0.64 | 8.46 ± 0.82 | 8.44 ± 0.61 | 6.10 ± 0.58 |
| FxT $\lambda$2 | 3.32 ± 0.27 | 3.92 ± 0.31 | 6.49 ± 0.56 | 6.06 ± 0.54 | 8.35 ± 0.54 | 9.00 ± 0.79 | 6.19 ± 0.53 |
| FxT $\lambda$3 | 3.34 ± 0.35 | 4.19 ± 0.38 | 6.55 ± 0.75 | 6.36 ± 0.81 | 8.36 ± 0.59 | 9.65 ± 1.28 | 6.41 ± 0.76 |

Table 10: SIB-200 Accuracy and Compression Results for with 3x Configurations

| Model | en | es | ru | hi | te | **ur** (OOD) | Avg |
|---|---|---|---|---|---|---|---|
| | | | **Accuracy** | | | | |
| BPE | 73.09 | 69.9 | 65.95 | 61.48 | 68 | 54.11 | 65.42 |
| FIXED | 73.25 | 69.84 | 66.47 | 61.59 | 67.52 | 57.09 | 65.96 |
| Binomial | 72.87 | 70.28 | 65.93 | 62.26 | 66.11 | 54.79 | 65.37 |
| FXT $\lambda 1$ | **73.51** | 70.22 | 66.47 | **62.42** | 67.11 | 56.99 | 66.12 |
| FXT $\lambda 2$ | 73.21 | **70.84** | **66.97** | 62.16 | 66.71 | **57.58** | **66.25** |
| FXT $\lambda 3$ | 73.35 | 70.22 | 66.75 | 62.36 | **67.82** | 57.33 | 66.31 |
| | | | **Compression Rate $\pm$ Std** | | | | |
| Binomial | 3.13 $\pm$ 0.30 | 3.79 $\pm$ 0.48 | 6.10 $\pm$ 0.74 | 9.85 $\pm$ 1.28 | 8.37 $\pm$ 1.21 | 8.58 $\pm$ 0.82 | 6.64 $\pm$ 0.88 |
| FXT $\lambda 1$ | 3.17 $\pm$ 0.19 | 3.89 $\pm$ 0.26 | 6.47 $\pm$ 0.53 | 7.99 $\pm$ 0.75 | 8.39 $\pm$ 0.58 | 8.52 $\pm$ 0.71 | 6.40 $\pm$ 0.55 |
| FXT $\lambda 2$ | 3.36 $\pm$ 0.26 | 4.10 $\pm$ 0.30 | 6.98 $\pm$ 0.60 | 9.18 $\pm$ 0.85 | 8.62 $\pm$ 0.65 | 8.73 $\pm$ 0.73 | 6.83 $\pm$ 0.60 |
| FXT $\lambda 3$ | 3.56 $\pm$ 0.31 | 4.32 $\pm$ 0.34 | 7.45 $\pm$ 0.72 | 10.06 $\pm$ 1.17 | 8.95 $\pm$ 0.74 | 9.07 $\pm$ 0.80 | 7.24 $\pm$ 0.74 |

Table 11: XNLI Accuracy and Compression Results for 3x Configurations

| Model | ILI (hi) | Med. Abs. (en) | Irony (en) |
|---|---|---|---|
| | | **Accuracy** | |
| BPE | 89.06 | 57.68 | 67.86 |
| FIXED | 89.03 | 60.18 | 65.68 |
| BINOMIAL | 89.47 | 62.81 | 67.60 |
| FXT $\lambda 1$ | 89.58 | 62.92 | 68.37 |
| FXT $\lambda 2$ | **90.33** | 62.74 | 68.75 |
| FXT $\lambda 3$ | 89.55 | **63.19** | **69.26** |
| | | **Compression Rate $\pm$ Std** | |
| Binomial 3x | 8.02 $\pm$ 1.38 | 3.01 $\pm$ 0.13 | 3.05 $\pm$ 0.14 |
| FXT $\lambda 1$ | 8.04 $\pm$ 0.89 | 3.11 $\pm$ 0.13 | 3.09 $\pm$ 0.08 |
| FXT $\lambda 2$ | 8.35 $\pm$ 0.87 | 3.21 $\pm$ 0.15 | 3.22 $\pm$ 0.31 |
| FXT $\lambda 3$ | **8.77** $\pm$ 1.21 | **3.43** $\pm$ 0.18 | **3.36** $\pm$ 0.13 |

Table 12: ILI, Medical Abstracts, and Irony (for 3$\times$ Configuration)

| MMLU and COPA | | | | | |
|---|---|---|---|---|---|
| **Model** | **copa (en)** | **copa (es)** | **mmlu (en)** | **mmlu (hi)** | |
| BPE | 55.00 | 49.20 | 23.50 | 24.30 | |
| BINOMIAL | 55.00 | 50.40 | 23.00 | 24.30 | |
| FxT $\lambda 3$ | **56.00** | **54.40** | **23.30** | **24.80** | |
| **INCLUDE** | | | | | |
| **Model** | **hi** | **ru** | **es** | **te** | **uk** |
| BPE | 25.05 | 26.99 | **27.45** | 24.09 | 33.64 |
| BINOMIAL | 23.77 | **25.18** | 28.00 | 23.72 | 34.55 |
| FxT $\lambda 3$ | 23.95 | 25.72 | 26.55 | 23.91 | **35.09** |

Table 13: Accuracy on MMLU, COPA, and INCLUDE tasks

| **Model** | **en** | **es** | **ru** | **uk** | **hi** | **te** | **Avg** |
|---|---|---|---|---|---|---|---|
| **Bits Per Byte (BPB)** | | | | | | | |
| Hnet | 1.38 | 1.27 | 0.77 | 0.77 | 0.60 | 0.56 | 0.89 |
| FxT | 1.38 | 1.27 | 0.77 | 0.76 | 0.60 | 0.56 | 0.89 |
| **Compression Rate $\pm$ Std** | | | | | | | |
| Hnet | $3.01 \pm 0.11$ | $3.58 \pm 0.13$ | $5.99 \pm 0.35$ | $5.60 \pm 0.25$ | $7.66 \pm 0.47$ | $8.01 \pm 4.03$ | $5.64 \pm 0.89$ |
| FxT | $3.43 \pm 0.16$ | $3.97 \pm 0.16$ | $6.68 \pm 0.33$ | $6.24 \pm 0.28$ | $8.55 \pm 0.52$ | $8.91 \pm 4.03$ | $6.30 \pm 0.90$ |

Table 14: Bits Per Byte (BPB) and Compression Results across Languages

| **Model** | **en-es** | **en-ru** | **en-hi** |
|---|---|---|---|
| BPE | 59.46 | 52.53 | 50.94 |
| BINOMIAL | 63.05 | 57.33 | 54.35 |
| FxT $\lambda 3$ | **64.08** | **57.76** | **54.73** |

Table 15: COMET scores on OPUS-100 machine translation task for our small-sized models. FLEXITOKENS outperforms across all languages.

| Model | ILI (hi) | Med. Abstract (en) |
|---|---|---|
| **Accuracy** | | |
| FxT 10x | 89.07 | 62.95 |
| FxT 5x | 89.28 | **63.47** |
| FxT 3x | **90.33** | 62.74 |
| **Compression Rate ± Std** | | |
| FxT 10x | 38.80 ± 16.75 | 13.22 ± 2.15 |
| FxT 5x | 14.82 ± 3.00 | 5.63 ± 0.33 |
| FxT 3x | 8.35 ± 0.87 | 3.21 ± 0.15 |

Table 16: ILI (hi) and Medical Abstract (en) $\lambda$ Ablation: Accuracy and Compression Results

| Model | en | es | ru | uk | hi | te | Avg |
|---|---|---|---|---|---|---|---|
| **Accuracy** | | | | | | | |
| FxT 10x | 57.35 | 59.80 | 55.88 | 50.98 | 47.06 | 51.47 | 53.76 |
| FxT 5x | **78.92** | **78.92** | 74.51 | **73.04** | 62.75 | 58.82 | 71.16 |
| FxT 3x | 77.94 | 75.98 | **74.51** | 71.57 | **69.12** | **66.18** | **72.55** |
| **Compression Rate ± Std** | | | | | | | |
| FxT 10x | 19.37 ± 8.23 | 16.23 ± 4.45 | 24.57 ± 6.82 | 28.69 ± 8.88 | 40.06 ± 14.68 | 44.43 ± 17.47 | 28.89 ± 11.06 |
| FxT 5x | 5.75 ± 0.65 | 6.78 ± 0.71 | 12.58 ± 1.91 | 10.62 ± 1.70 | 13.42 ± 1.63 | 15.17 ± 2.04 | 10.72 ± 1.54 |
| FxT 3x | 3.32 ± 0.27 | 3.92 ± 0.31 | 6.49 ± 0.56 | 6.06 ± 0.54 | 8.35 ± 0.54 | 9.00 ± 0.79 | 6.19 ± 0.53 |

Table 17: SIB-200 $\alpha$ Ablation: Accuracy and Compression Results

| Model | en | es | ru | uk | hi | te | Avg |
|---|---|---|---|---|---|---|---|
| **F1 Score** | | | | | | | |
| FxT 10x | 61.81 | 75.48 | 66.90 | 76.90 | 59.88 | 45.15 | 64.35 |
| FxT 5x | 62.84 | 75.81 | 67.48 | 77.68 | 60.02 | 45.66 | 64.92 |
| FxT 3x | **63.96** | **76.23** | **67.55** | **77.99** | **62.24** | **48.13** | **66.02** |
| **Compression Rate ± Std** | | | | | | | |
| FxT 10x | 14.15 ± 6.07 | 16.87 ± 6.39 | 40.03 ± 19.10 | 27.91 ± 11.95 | 42.52 ± 21.82 | 26.55 ± 11.73 | 28.01 ± 14.14 |
| FxT 5x | 5.83 ± 1.23 | 7.26 ± 2.01 | 15.30 ± 5.90 | 11.93 ± 3.59 | 15.92 ± 4.68 | 10.80 ± 2.57 | 11.17 ± 3.69 |
| FxT 3x | 3.27 ± 0.44 | 3.93 ± 0.58 | 8.52 ± 1.49 | 6.58 ± 1.38 | 9.15 ± 2.21 | 6.12 ± 1.00 | 6.26 ± 1.33 |

Table 18: WikiANN $\alpha$ Ablation: F1 Score and Compression Results

| Model | en | es | ru | hi | te | ur | Avg |
|---|---|---|---|---|---|---|---|
| **Accuracy** | | | | | | | |
| FxT 10x | 71.42 | 68.60 | 65.59 | 62.22 | 66.05 | 57.52 | 65.23 |
| FxT 5x | 72.97 | 70.38 | 65.47 | 61.88 | 65.49 | 56.71 | 65.48 |
| FxT 3x | **73.21** | **70.84** | **66.97** | **62.16** | **66.71** | **57.58** | **66.25** |
| **Compression Rate ± Std** | | | | | | | |
| FxT 10x | 13.41 ± 2.88 | 15.88 ± 3.12 | 25.20 ± 6.07 | 41.81 ± 12.06 | 37.23 ± 8.77 | 40.84 ± 12.71 | 29.06 ± 8.55 |
| FxT 5x | 6.06 ± 0.72 | 7.59 ± 0.88 | 13.02 ± 2.08 | 15.44 ± 2.16 | 15.10 ± 1.60 | 15.67 ± 2.40 | 12.15 ± 1.76 |
| FxT 3x | 3.36 ± 0.26 | 4.10 ± 0.30 | 6.98 ± 0.60 | 9.18 ± 0.85 | 8.62 ± 0.65 | 8.73 ± 0.73 | 6.83 ± 0.60 |

Table 19: XNLI $\alpha$ Ablation: Accuracy and Compression Results

| Model | es | ru | hi | avg |
|---|---|---|---|---|
| **Accuracy** | | | | |
| FxT 10x | 77.67 | 87.07 | **54.24** | **72.99** |
| FxT 5x | 77.74 | 87.17 | 52.71 | 72.54 |
| FxT 3x | **77.77** | **87.33** | 53.12 | 72.74 |
| **Compression Rate ± Std** | | | | |
| FxT 10x | 16.07 ± 4.53 | 26.55 ± 8.33 | 39.60 ± 21.99 | 27.41 ± 12.12 |
| FxT 5x | 6.83 ± 0.77 | 11.45 ± 2.11 | 15.47 ± 5.21 | 11.25 ± 2.86 |
| FxT 3x | 3.90 ± 0.28 | 6.44 ± 0.61 | 8.16 ± 1.65 | 6.17 ± 1.03 |

Table 20: Multilingual Sentiment $\alpha$ Ablation: Accuracy and Compression Results

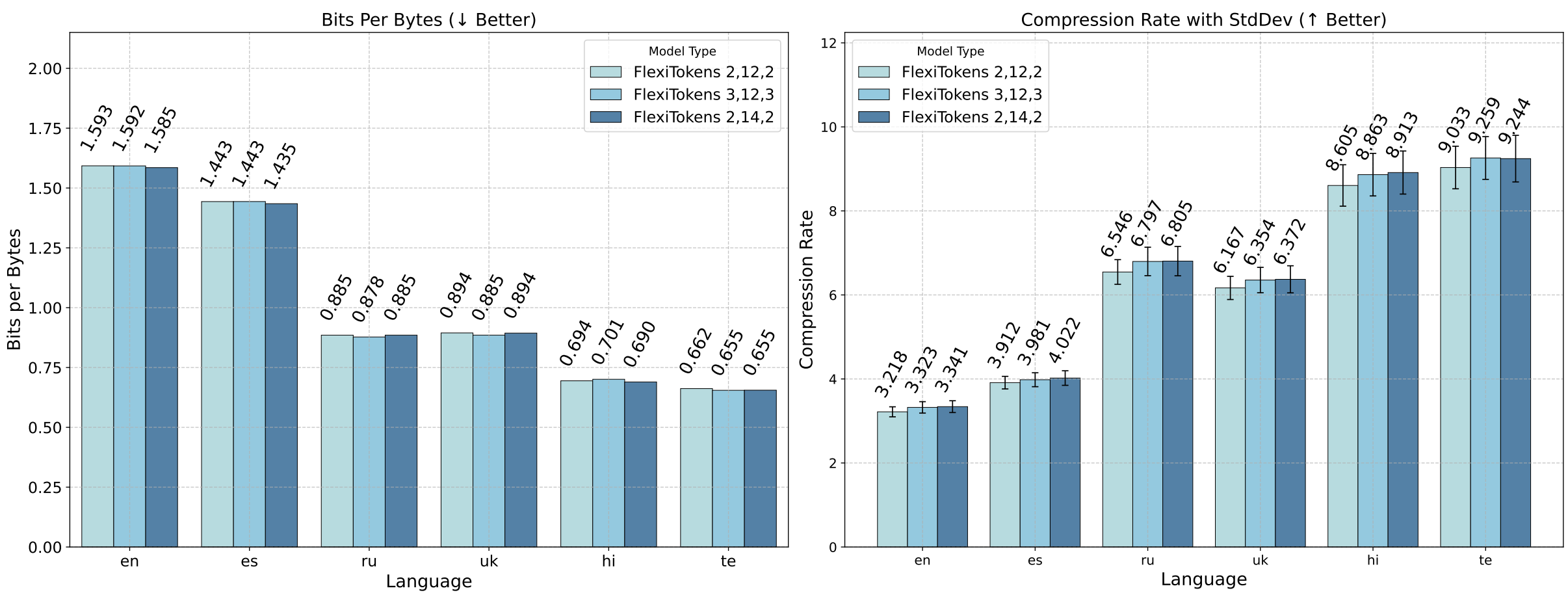

Figure 7: FineWeb Test results for ablating the number layers in FLEXITOKENS. Adding more layers results to lower BPB and higher compression rate across all model sizes. FLEXITOKENS (2,12,2) is equivalent to 2, 12 and 2 transformer layers in the tokenization, LM, and upsampling module, respectively.

<table>
<tr><th>Tokenizer</th><th>Sentence and Segmentation</th><th>#Tokens</th></tr>
<tr><td>ur</td><td>39-year-old SpongeBob was diagnosed with hypertrophic cardiomyopathy in Mumbai.</td><td>–</td></tr>
<tr><td>BPE</td><td>39|Ø|³|Ø§ÙĦ|Û|ģ| |Ø§|Ø|³|Ù¾|ÙĪ|Ø|¬| Ø|¨|Ø§|Ø|¨|<br>|Ú©|ÙĪ| Ùħ|Ùħ|Ø|¨|Ø|¦|Û|Į| Ùħ|Û|Į|Úº|<br>|Û|ģ|Ø§|Ø|¦|Ù¾|Ø±|Ù¹|Ø±|Ø§|Ù|ģ|Ú©|<br>|Ú©|Ø§|Ø±|ÚĪ|Û|Į|ÙĪ|Ùħ|Û|Į|ÙĪ|Ù¾|Û|Į|Ø|ª|Ú¾|Û|Į|<br>|Ú©|Û|Į| Ø|ª|Ø|´|Ø|®|Û|Į|Ø|µ| |Û|ģ|ÙĪ|Ø|¦|Û|Į|Û|Ķ</td><td>107</td></tr>
<tr><td>BINOMIAL 3×</td><td>3|9 |سال |اسپنج |باب |کو |ممبئ| |میں |ہائپر|ٹر|افک |کارڈیو|میو|پیتھی | کی<br>|تشخ|ص |ہو|ئی۔</td><td>21</td></tr>
<tr><td>FLEXITOKENS 3×</td><td>93 |سال |اسپنج |باب |کو |ممبئی |میں |ہائپر|ٹر|افک |کارڈیو|میو|پ|تھ| | کی<br>|تشخیص |ہوئی۔</td><td>17</td></tr>
<tr><td>te</td><td>He spent the whole night watching Netflix. He fell asleep early.</td><td>–</td></tr>
<tr><td>BPE</td><td>à°ħà°¤|à°¡|à±ģ| à°°|à°¾à°¤|à±į|à°°|à°Ĥà°¤|à°¾|<br>à°¨|à±Ĩà°Ł|à±į|âĢ|Į|à°«|à±į|à°²|à°¿à°ķ|à±į|à°¸|à±į| à°ļ|à±Ĥà°¸|à±į|à°¤|à±Ĥ|<br>à°Ĺà°¡|à°¿à°ª|à°¾à°¡|à±ģ. à°ħà°¤à°¡|à±ģ| à°¤|à±į|à°µà°°à°Ĺ|à°¾|<br>à°¨|à°¿à°¦|à±į|à°°à°ª|à±ĭà°¯|à°¾à°¡|à±ģ.</td><td>37</td></tr>
<tr><td>BINOMIAL 3×</td><td>అ|డ| రా|రంౌ నె్ఫ|లి|స్ చూస్తు| గడి|ఌ. అతడ| త|వరగ| నీ్రపౌ|ఌ.</td><td>22</td></tr>
<tr><td>FLEXITOKENS 3×</td><td>అతడ| రాత|రంతా |నెట్ఫ|లి|క|స్ |చూస్తు| గడిపాడ:. అతడ| త్వరగౌ<br>|నిద|రప|యాడ:.</td><td>17</td></tr>
<tr><td>en</td><td>Influenza and pneumonia were identified as major causes of mortality in children.</td><td>–</td></tr>
<tr><td>BPE</td><td>In|f|lu|enza| and| p|ne|um|onia| were| identified|<br>as| major| causes| of| mort|ality| in| children|.</td><td>20</td></tr>
<tr><td>Binomial 3×</td><td>I|n|fl|uenza |an|d |pn|eumon|ia |wer|e<br>|id|ent|ified |as |maj|or| |causes |of<br>|mor|t|ality |in| |chil|dren.</td><td>25</td></tr>
<tr><td>FLEXITOKENS 3×</td><td>Infl|uenz|a |and| |pneu|m|onia |were |identified<br>|as |m|ajor| |causes |of |m|or|tality |in<br>|childr|en.</td><td>20</td></tr>
<tr><td>en</td><td>Oh no, another surprise bonus at work. Just what I didn’t need 😀😂😂🙋🏽‍♂️🤸‍♀️.</td><td>–</td></tr>
<tr><td>BPE</td><td>Oh| no|,| another| surprise| bonus| at| work|.| Just| what| I| didn|âĢĻt| need|<br>ðŁĺ|Ģ|ðŁĺ|Ĥ|ðŁĺ|Ĥ|ðŁ|Ļ|ĭ|ðŁı|½|âĢ|į|âĻ|Ĥ|ï¸ı|ðŁ|¤¸|ðŁı|¾|âĢ|<br>į|âĻ|Ģ|ï¸ı|ł|ħ|§|ł|ħ|ķ|ł|Ħ|Ĳ|ł|ħ|ĳ|ł|ħ|¢|ł|ħ|ķ|ł|Ħ|Ĳ|ł|ħ|ĵ|ł|ħ|Ł|ł|ħ|Ł|ł|ħ|Ľ|ł|ħ|ķ|ł|ħ|Ķ|.</td><td>82</td></tr>
<tr><td>Binomial 3×</td><td>Oh no, |an|ot|her| |sur|pr|ise |bon|us |at|<br>|w|ork. |Just| |w|hat |I didn’|t| |need |<br>😀|||||😂😂| 🙋🏽‍♂️||||🤸‍♀️|.</td><td>33</td></tr>
<tr><td>FLEXITOKENS 3×</td><td>Oh |no, |anot|her |sur|pr|ise |bonus |at| |wor|k.|<br>|J|ust |what| |I |didn|t| |need | 😀😂😂| 🙋🏽‍♂️|<br>🤸‍♀️|.</td><td>25</td></tr>
</table>

Table 21: Tokenization outputs with different methods (Urdu, Telugu, English)